\documentclass[9pt,twocolumn,a4paper,conference]{IEEEtran} 

\usepackage[printonlyused]{acronym}
\usepackage{amssymb,amsmath,amsfonts}
\usepackage{ifthen}
\usepackage{hyperref}
\usepackage{url}
\usepackage{pstricks}
\usepackage{graphicx}
\usepackage{subcaption}

\newcommand{\real}{\mathbb{R}}
\newcommand{\ma}[1]{{\mathbf #1}}

\DeclareMathOperator*{\argmin}{argmin}	
\DeclareMathOperator*{\argmax}{argmax}

\newcommand{\mbf}{\mathbf}

\newcommand{\bm}{\boldsymbol}
\renewcommand{\epsilon}{\varepsilon}

\newtheorem{theorem}{Theorem}

\newtheorem{proposition}{Proposition}

\newtheorem{remark}{Remark}

\newcommand{\vertfill}[1]{\rotatebox{90}{#1}}

\usepackage{lipsum}
\usepackage{xparse}
\newsavebox{\fminipagebox}
\NewDocumentEnvironment{fminipage}{m O{\fboxsep}}
 {\par\kern#2\noindent\begin{lrbox}{\fminipagebox}
  \begin{minipage}{#1}\ignorespaces}
 {\end{minipage}\end{lrbox}%
  \makebox[#1]{%
    \kern\dimexpr-\fboxsep-\fboxrule\relax
    \fbox{\usebox{\fminipagebox}}%
    \kern\dimexpr-\fboxsep-\fboxrule\relax
  }\par\kern#2
 }

\title{On Generation of Adversarial Examples \\ using Convex Programming}

\author{Emilio Rafael Balda, Arash Behboodi, Rudolf Mathar \\
 \texttt{\{emilio.balda, arash.behboodi, mathar\}@ti.rwth-aachen.de}\\
 \footnotesize{Institute for Theoretical Information Technology, RWTH Aachen University,  Aachen, 52074}
}
\date{}
\begin{document}
\acrodef{DNN}{Deep Neural Network}
\acrodef{FGSM}{Fast Gradient Sign Method}
\acrodef{BALDA}{Bounded Attack with Linearly Directed Approximation}
\acrodef{BIM}{Basic Iterative Method}
\acrodef{PGD}{Projected Gradient Descent}
\acrodef{FCNN}{Fully Connected Neural Network}

\maketitle

\begin{abstract}
It has been observed that deep learning architectures tend to make erroneous decisions with high reliability for particularly designed adversarial instances. 
In this work, we show that the perturbation analysis of these architectures provides a framework for generating adversarial instances by convex programming which, for classification tasks, is able to recover variants of existing non-adaptive adversarial methods. The proposed framework can be used for the design of adversarial noise under various desirable constraints and different types of networks. Moreover, this framework is capable of explaining various existing adversarial methods and can be used to derive new algorithms as well. We make use of these results to obtain novel algorithms. The experiments show the competitive performance of the obtained solutions, in terms of fooling ratio, when benchmarked with well-known adversarial methods.\footnote{For the sake of reproducible research, the tensorflow implementations used in this paper have been made available at \mbox{\url{hppts://github.com/ebalda/adversarialconvex}}.}
\end{abstract}
\section{Introduction}
\label{sec:intro}
%
\acp{DNN} enjoy excellent performances in speech analysis \cite{hinton_deep_2012} and visual tasks \cite{krizhevsky_imagenet_2012,he_deep_2016,szegedy_going_2015,ren_faster_2017}. 
Despite their success, they have been shown to suffer from instability in their classification under adversarial perturbations \cite{szegedy_intriguing_2014}. 
Adversarial perturbations are intentionally worst-case designed noise that aims at changing the output of a \ac{DNN} to an incorrect one. 
Interestingly, the example of adversarial perturbations on the ImageNet dataset show that adversarial examples are almost indistinguishable to the human eye from the original images.  
Related to the concept of adversarial perturbations is the notion of rubbish class or fooling images \cite{DNNareEasilyFooled,goodfellow_explaining_2014} where the examples are clearly perceived by the human eye as not belonging to any categories in the training set but nevertheless classified with high confidence as one of the categories by \acp{DNN}. 
Moreover, as in \cite{goodfellow_explaining_2014,DeepFool,UAP}, the adversarial method has access to the input of a neural network based system. 
In this context, the attacker would attempt to apply perturbations to system input that are not perceived by the system's administrator, such that the performance of the system is severely degraded. 

%

This is of significant importance in safety critical systems such as autonomous driving architectures and surveillance applications. 
These discoveries gave rise to extensive research on understanding the instability of \acp{DNN} (for instance refer to \cite{akhtar_threat_2018,wang_theoretical_2017,fawzi_fundamental_2015} and references therein). 
Although \acp{DNN} might achieve robustness to random noise \cite{fawzi_robustness_2016}, it has been shown that there is a clear distinction between the robustness of a classifier to random noise and its robustness to adversarial perturbations. 
In \cite{szegedy_intriguing_2014}, the adversarial perturbation was obtained to  maximize the prediction error at the output and it was approximated using box-constrained L-BFGS. 
The \ac{FGSM} in \cite{goodfellow_explaining_2014} was based on finding the scaled sign of the gradient of the cost function. 
Note that the \ac{FGSM} aims at minimizing $\ell_\infty$-norm of the perturbation while the former algorithm minimizes $\ell_2$-norm of the perturbation under box constraint on the perturbed example. 
In practice, the perturbed input values are limited inside a certain dynamic range, such as values between 0 and 1 for the case of images. 
The algorithm DeepFool \cite{DeepFool} utilizes an iterative linearization of the \ac{DNN} to generate perturbations that are minimal in $\ell_p$-norm for $p>1$. Although improving on the \ac{FGSM}, the algorithm is an iterative method requiring calculation of the gradient function at each step and its convergence rate, which depends on a previously selected step size parameter, is not guaranteed.  
In \cite{BIM2016} the authors propose an iterative version of the \ac{FGSM}, called \ac{BIM}.
This method was later extended in \cite{pgd_attack}, where randomness was introduced in the computation of adversarial perturbations. 
This attack is called the \ac{PGD} method. 
An interesting feature of these algorithms is that some of the perturbations generalize over other datasets and \acp{DNN} \cite{UAP,goodfellow_explaining_2014}. 
These perturbations are called universal adversarial perturbations. 
This is partly explained by the fact that certain underlying properties of the perturbation, such as direction in case of image perturbation, matters the most and therefore generalized through different datasets.

There are various theories regarding the nature of adversarial examples. The authors in \cite{goodfellow_explaining_2014} propose the linearity hypothesis where the existence of adversarial images is attributed to the approximate linearity  of classifiers, although this hypothesis has been challenged in \cite{tanay_boundary_2016}. 
There are other theories focusing mostly on decision boundaries of classifiers and their analytic properties \cite{fawzi_robustness_2016,fawzi_robustness_2017}. 

In this paper, the adversarial examples are generated using an approximation of the target classifier with an affine function.
In Section \ref{sec:pertclass}, we first introduce the concept of a first-order perturbation analysis, and its application to neural network classifiers. 
Then, in Section \ref{sec:avd_and_rub}, the first-order perturbation analysis is utilized to formulate the general formula for generation of adversarial attacks as a convex optimization problem. 
In particular it is shown that the worst perturbation incurred by an imperceptible adversarial perturbation can be found using a convex optimization problem and the closed form solutions  are provided for the classification problem. In Section \ref{sec:other}, we show the applicability of our framework for various learning tasks such as regression, image segmentation, and detection.
Later, in Section \ref{sec:expres}, we benchmark the obtained methods, for the context of image classification, against the \ac{FGSM} and DeepFool algorithms. In addition, we show that these algorithms can be formulated within our framework. 
Furthermore, it is shown that our proposed algorithm manages to outperform existing methods using empirical simulations {on the MNIST and CIFAR-10 datasets}. 

\section{Perturbation Analysis of General Classifiers} \label{sec:pertclass}

The perturbation analysis, also called sensitivity analysis, is used in signal processing for analytically quantifying the error at the output of a system that occurs as consequence of a known perturbation at the system's input. Adversarial images can also be considered as a slightly perturbed version of original images that manage to change the output of the classifier. 
Indeed, the adversarial methods in \cite{DeepFool,goodfellow_explaining_2014} are implicitly based on approximating the effect of an input perturbation on a relevant function which is either the classifier function or the cost function used for training. 
The perturbation analysis of classifiers provide a unifying view of previous methods.

For a classifier given as a function $f(.)$ of inputs $\ma x$, if the input vector is perturbed by a sufficiently small perturbation $\Delta \ma{x}^{(0)}$ given by $\hat{\ma{x}}^{(0)} = \ma{x}^{(0)} + \Delta \ma{x}^{(0)}$, the first-order perturbation incurred at the output is given by the first-order Taylor series of $f(\ma{x}^{(0)} + \Delta \ma{x}^{(0)})$ as
\begin{equation} \label{eq:Taylor}
f(\ma{x}^{(0)} + \Delta \ma{x}^{(0)}) \approx f(\ma{x}^{(0)}) + \ma{J}_f(\ma{x}^{(0)})  \Delta \ma{x}^{(0)}
\end{equation}
where $\ma{J}_f(\ma{x})$ is the Jacobian of the function $f(\ma{x})$. Therefore, the error at the output of the classifier can be approximated as $f(\ma{x}^{(0)} + \Delta \ma{x}^{(0)}) - f(\ma{x}^{(0)}) \approx \ma{J}_f(\ma{x}^{(0)})  \Delta \ma{x}^{(0)}$. For neural networks, the perturbation analysis has been previously studied as in \cite{sensitivity}. 

Consider an $L$-layered neural network with the input vector $\ma{x}^{(0)} \in \real^{m_0}$ and the corresponding  output vector $\ma{x}^{(L)} \in \real^{m_L}$ , layer sizes $(m_1,\dots,m_{L})$, the weights of the hidden layer $l$ denoted by the matrix $\ma{W}^{(l)} \in \real^{m_{l} \times m_{l-1}}$, the bias vector  by $\ma{b}^{(l)} \in \real^{m_l}$, and differentiable point-wise activation functions by $\phi^{(l)}$ with the derivative ${\phi^{(l)}}'$ at the layer $l\in[L]$\footnote{In this work $[L]\triangleq\{1,\dots,L\}$.}. 
Let the function $f: \real^{m_0} \rightarrow \real^{m_L}$ be the \ac{DNN}'s function that maps the input vector $\ma{x}^{(0)}$ to the output vector $\ma{x}^{(L)}$. The following proposition provides a first-order perturbation analysis of \acp{DNN}.
\begin{theorem}
For a given $\ma x^{(0)}$, the first-order perturbation at the output of a \ac{DNN}, $\Delta f$, with continuously differentiable activation functions caused by a small input perturbation $\Delta\ma x^{(0)}$ is given by:
\[
 \Delta f=\ma{Z}^{(L)}\Delta\ma x^{(0)},
\]
where $\ma{Z}^{(L)}$ is the Jacobian of the \ac{DNN} function $\ma{J}_f(\ma{x}^{(0)})$ given by:
\[
 \ma{Z}^{(L)} = \ma{D}^{(L)} \cdot \ma{W}^{(L)} \cdot \ma{D}^{(l-1)} \cdot \ma{W}^{(l-1)} \cdots \ma{D}^{(1)} \cdot \ma{W}^{(1)} \, ,
\]
with $\ma{D}^{(l)} \triangleq \mathrm{Diag}\{ {\phi^{(l)}}'( \ma{W}^{(l)} \cdot \ma{x}^{(l-1)} + \ma{b}^{(l)} ) \}$.
\label{thm:pert}
\end{theorem}

The proof is based on approximating each layer with a linear function. The output perturbation follows from consecutive application of linear approximations. In the $L$-layered neural network, the $l$-th layer output  $\ma{x}^{(l)}$ is given as 
\begin{align*}
	\ma{x}^{(l)} &= \phi^{(l)} \left( \ma{W}^{(l)}   \ma{x}^{(l-1)} + \ma{b}^{(l)} \right) \quad \forall l\in[L] \, .
\end{align*}
In the context of perturbation analysis, it is assumed that all the system parameters (i.e., $\ma{x}^{(0)}$, $\phi^{(l)}$, $\ma{W}^{(l)}$, and $\ma{b}^{(l)}$) are known for $l\in[L]$.  

Suppose that there is a perturbation $\Delta \ma{x}^{(l-1)}$ at the $l$-th layer output and the perturbed version of the $l$-th layer output be given by $\hat{\ma{x}}^{(l)} = \ma{x}^{(l)} + \Delta \ma{x}^{(l)}$. From the above relations we have
\begin{align*}
\hat{\ma{x}}^{(l)} &= \phi^{(l)} \left( \ma{W}^{(l)} \hat{\ma{x}}^{(l-1)} + \ma{b}^{(l)} \right)\\
&= \phi^{(l)} \left( \ma{W}^{(l)} \ma{x}^{(l-1)} +  \ma{W}^{(l)} \Delta\ma{x}^{(l-1)}+ \ma{b}^{(l)} \right)\\
&\approx \ma{x}^{(l)}+\mathbf J_{\phi^{(l)}}\left(\ma{W}^{(l)}   \ma{x}^{(l-1)} + \ma{b}^{(l)}\right)\ma{W}^{(l)} \Delta\ma{x}^{(l-1)}.
\end{align*}
But since $l$-th layer activation function is applied in a point-wise fashion, its Jacobian is given by a diagonal matrix that is
\[
 \ma{D}^{(l)} \triangleq \mathrm{Diag}\{ {\phi^{(l)}}'( \ma{W}^{(l)} \cdot \ma{x}^{(l-1)} + \ma{b}^{(l)} ) \}.
\]
Therefore the perturbation of $l$-th layer is given by:
\[
 \Delta\ma{x}^{(l)}=\ma{D}^{(l)}  \ma{W}^{(l)} \Delta\ma{x}^{(l-1)}.
\]
By consecutive application of this result, $\Delta \ma{x}^{(l)}$ can be approximated as $\Delta \ma{x}^{(l)} \approx \ma{Z}^{(l)} \cdot \Delta \ma{x}^{(0)}$, where
\[
\ma{Z}^{(l)} = \ma{D}^{(l)} \cdot \ma{W}^{(l)} \cdot \ma{D}^{(l-1)} \cdot \ma{W}^{(l-1)} \cdots \ma{D}^{(1)} \cdot \ma{W}^{(1)} \, .
\]

The above theorem can be applied to general classifiers as well as other learning functions such as regression. 
Note that if the activation functions are not differentiable at some points, one can instead use sub-derivatives instead. 
One can recourse to higher order perturbation analysis where the perturbation is quadratic or higher order function of $\Delta\mathbf x^{(0)}$. 
This might be necessary if the perturbation affects the output mainly through its higher orders, for example when the perturbation belongs to the null space of the Jacobian matrix.


\section{Generating Malicious Examples via Convex Programming} \label{sec:avd_and_rub}
As mentioned before, the adversarial examples can be considered as perturbed versions of training examples and hence the analysis above fits our scenario, where the adversarial perturbation $\Delta \ma{x}^{(0)}$ should be imperceptible to the target system. In \cite{goodfellow_explaining_2014}, the proposed method is based on finding a perturbation with bounded $\ell_\infty$-norm that maximizes the error function used for the training which utilizes the first-order perturbation analysis of the error function. On the other hand in \cite{DeepFool}, the authors directly minimize the norm of the perturbation that changes the classifier's output. Their analysis is based on linearized approximation of the underlying classifier which is indeed its first order-perturbation analysis. While \mbox{DeepFool} might generate perturbations that are perceptible, the \mbox{\ac{FGSM}} might not change the classifier output. In this work, we proposed another method based on the first-order perturbation analysis that targets the classifier's output directly and simultaneously guarantees that the perturbation is imperceptible.  

For classification tasks, let $k: \real^{m_0} \rightarrow \{1,2,\dots,m_L\}$ be the classifier function that maps the input $\mathbf{x} \in \real^{m_0}$ to its estimated label $k\left(\mathbf{x}\right) \in \{1,2,\dots,m_L\}$. The function $k$, defined in this way, is not differentiable anymore. However. In the context of classification, there is a proxy function $f(\mathbf{x})$ given by a vector $(f_1(\mathbf{x}),\dots,f_m(\mathbf{x}))$ which has unit $\ell_1$-norm and with each of $m_L$ scalar functions  $f_l(\mathbf{x})$ interpreted as the probability of class belonging. The classifier $k$ is given then as
\begin{equation} \label{eq:classifier}
k(\mathbf{x}) = \argmax_{l \in[m_L] } \left\lbrace f_l \left(\mathbf{x} \right)  \right\rbrace \,.
\end{equation}
The input perturbation aims at changing the output of the classifier.  Suppose that the input vector is perturbed by a small perturbation $\bm{\eta} \in \real^{m_0}$.
Then, the classifier $k$ is said to be fooled by the adversarial sample $\hat{\mathbf{x}} = \mathbf{x} + \bm{\eta}$ if $k(\mathbf{x}) \neq k(\hat{\mathbf{x}})$, that is:
\begin{equation}\label{eq:L}
   L(\mathbf{x} + \bm{\eta}) = \min_{l \neq k(\mathbf{x})} \{ f_{k(\mathbf{x})}(\mathbf{x} + \bm{\eta}) - f_l(\mathbf{x} + \bm{\eta})  \} <0\, .
\end{equation}
However what is particularly disturbing in adversarial images is that the image looks almost unchanged to the naked eye. Therefore the input perturbation should not change the output the ground truth classifier, also called oracle classifier in \cite{wang_theoretical_2017}, which is here the naked eye. As in \cite{wang_theoretical_2017}, the proxy functions of the oracle classifier are denoted by $g_l$ and we should have:
\[
    L_o(\mathbf{x} + \bm{\eta}) = \min_{l \neq k(\mathbf{x})} \{ g_{k(\mathbf{x})}(\mathbf{x} + \bm{\eta}) - g_l(\mathbf{x} + \bm{\eta})  \} >0\, .
\]
Therefore the problem of adversarial design can be formulated as:
\begin{align}
 \text{Find:}& \quad \bm\eta\nonumber\\
 \text{s.t.} &\quad  L(\mathbf{x} + \bm{\eta})< 0, \quad L_o(\mathbf{x} + \bm{\eta})>0.
 \label{eq:originalOP}
\end{align}
There are two problems with the above formulation. First, the oracle function is not known in general and second the function $L$ can be non-convex. One solution is to linearize $L$ through the perturbation analysis performed on each individual function and replacing the constraint on the oracle function with a simpler one like $\ell_p$-norm of the perturbation.
%
%
%
%

The first order perturbation analysis of $L$ yields: 
\[L(\mathbf{x} + \bm{\eta}) \approx L(\mathbf{x}) + \bm{\eta}^{\mathrm{T}} \nabla_{\mathbf{x}} L(\mathbf{x}),
 \]
where $\nabla_{\mathbf{x}} L(\mathbf{x})$ is the gradient of $L(\mathbf{x})$.  The condition that corresponds to the oracle function can be approximated by
$\| \bm{\eta} \|_{p} \leq \epsilon$ for  sufficiently small $\epsilon \in \real^{+}$. This means that the noise is sufficiently small in $\ell_p$-norm sense so that the observer does not notice it. These Gradient and norm relaxations yield to the following alternative optimization problem:
\begin{align}
 \text{Find:}& \quad \bm\eta\nonumber\\
 \text{s.t.} &\quad  L(\mathbf{x}) + \bm{\eta}^{\mathrm{T}} \nabla_{\mathbf{x}} L(\mathbf{x})<0, \quad \| \bm{\eta} \|_{p} \leq \epsilon.
\tag{GN} 
 \label{eq:originalOPII}
\end{align}
The above problem was also derived in \cite{robustnessguarantees2017} and is a convex optimization problem that can be efficiently solved. As we will see later, this formulation of the problem can be relaxed into some well known existing adversarial methods. However it is interesting to observe that this problem is not always feasible as stated in the following proposition. 
\begin{proposition}
 The optimization problem \eqref{eq:originalOPII} is not feasible if for $q=\frac{p}{p-1}$
 \begin{equation}
  \epsilon \|\nabla_{\mathbf{x}} L(\mathbf{x})\|_q < L(\mathbf x).
  \label{eq:condition}
 \end{equation}
 \label{prop:feasibility}
\end{proposition}
\textbf{Proof.} The proof follows a simple duality argument and is an elementary optimization theory result. We repeat the proof for completeness. Note that the dual norm of $\ell_p$ is defined by:
\[
 \|\ma x\|_p^*=\sup\{\ma a^T\ma x: \|\ma a\|_p\leq 1\}.
\]
 Furthermore $ \|\ma x\|_p^*= \|\ma x\|_q$ for $q=\frac{p}{p-1}$.
Since the $\ell_p$-norm of $\bm\eta$ is bounded by $\epsilon$, the value of $\bm{\eta}^{\mathrm{T}} \nabla_{\mathbf{x}} L(\mathbf{x})$ is always bigger than $-\epsilon \|\nabla_{\mathbf{x}} L(\mathbf{x})\|_p^*$. However if the condition \ref{eq:condition} holds, then we have:
\[
 L(\mathbf{x}) + \bm{\eta}^{\mathrm{T}} \nabla_{\mathbf{x}} L(\mathbf{x})\geq L(\mathbf{x}) -\epsilon \|\nabla_{\mathbf{x}} L(\mathbf{x})\|_p^*>0.
\]
Therefore, the problem is not feasible. $\blacksquare$

%
Proposition \ref{prop:feasibility} shows that given a vector $\mathbf x$, the adversarial perturbation should have at least $\ell_p$-norm equal to $\frac{ L(\mathbf x)  }{\|\nabla_{\mathbf{x}} L(\mathbf{x})\|_q}$. In other words if the ratio $\frac{ L(\mathbf x)  }{\|\nabla_{\mathbf{x}} L(\mathbf{x})\|_q}$ is too small, then it is easier to fool the network. In that sense,  Proposition \ref{prop:feasibility} provides an insight into the stability of classifiers. In \cite{DeepFool}, the authors suggest that the robustness of the classifiers can be measured as:
\[
 \hat{\rho}_{1}(f)=\frac{1}{|\mathcal D|}\sum_{\mathbf x\in\mathcal D}\frac{\|\hat{\mathbf r}(\mathbf x)\|_p}{\|\mathbf x\|_p},
\]
where $\mathcal D$ denotes the test set and $\hat{\mathbf r}(\mathbf x)$ is the minimum perturbation required to change the classifier's output. The above theorem suggests that one can also use the following as the measure of robustness:
\[
 \hat{\rho}_{2}(f)=\frac{1}{|\mathcal D|}\sum_{\mathbf x\in\mathcal D}\frac{ L(\mathbf x)  }{\|\nabla_{\mathbf{x}} L(\mathbf{x})\|_q}.
 \]
 The lower $ \hat{\rho}_{2}(f)$, the easier it gets to fool the classifier and therefore it becomes less robust to adversarial examples.  One can also look at other statistics related to $\frac{ L(\mathbf x)  }{\|\nabla_{\mathbf{x}} L(\mathbf{x})\|_q}$ in order to evaluate the robustness of classifiers.
 
Since Proposition \ref{prop:feasibility} shows that the optimization problem \eqref{eq:originalOPII} might not be feasible, alternatively we propose to solve the following optimization problem, called the Gradient-base Norm-constrained method:
\begin{equation}\label{eq:MainOpt}
\min_{\bm{\eta}} \left \{ L(\mathbf{x}) + \bm{\eta}^{\mathrm{T}} \nabla_{\mathbf{x}} L(\mathbf{x}) \right \} \quad \mathrm{s.t.} \quad \| \bm{\eta} \|_{p} \leq \epsilon \, ,
\tag{GNII}
\end{equation}
which finds the best perturbation under a given constraint. The constraint aims at guaranteeing that the adversarial images is still imperceptible by an ordinary observer. Note that \eqref{eq:MainOpt} is fundamentally different from \cite{DeepFool, robustnessguarantees2017}, where the norm of the noise does not appear as a constraint. Using a similar duality argument, the problem \eqref{eq:MainOpt} has a closed form solution given below.
\begin{proposition}
  If $\nabla_{\mathbf{x}} L(\mathbf{x})=\left(\frac{\partial L(\mathbf x)}{\partial x_1},\dots,\frac{\partial L(\mathbf x)}{\partial x_{m_0}}\right)$, the closed form solution to the problem \eqref{eq:MainOpt} is given by
\begin{align}
   & \bm{\eta} = - \epsilon \frac{ 1}{\|\nabla_{\mathbf{x}} L(\mathbf{x})\|_q^{q-1}}\times\nonumber \\
 &   \left( \mathrm{sign}(\frac{\partial L(\mathbf x)}{\partial x_1})\left|\frac{\partial L(\mathbf x)}{\partial x_1}\right|^{q-1} , \dots, \mathrm{sign}(\frac{\partial L(\mathbf x)}{\partial x_{m_0}})\left|\frac{\partial L(\mathbf x)}{\partial x_{m_0}}\right|^{q-1}\right)
 \label{eq:minMainOP}
\end{align}
for $q=\frac{p}{p-1}$. Particularly for $p=\infty$, we have $q=1$ and the solution is given by the following:
\begin{equation}
    \bm{\eta} = - \epsilon \, \mathrm{sign}(\nabla_{\mathbf{x}} L(\mathbf{x}) ) \, .
\end{equation}
\end{proposition}
\textbf{Proof.} Based on the duality definition, we know that 
\[
 \sup_{\|\bm \eta\|_p\leq 1}\bm{\eta}^{\mathrm{T}} \nabla_{\mathbf{x}} L(\mathbf{x})=\|\nabla_{\mathbf{x}} L(\mathbf{x})\|_p^*,
\]
which in turn implies that the objective function is lower bounded by $L(\mathbf{x}) -\epsilon \|\nabla_{\mathbf{x}} L(\mathbf{x})\|_p^*$. It is easy to verify that the minimum is attained by the expression \ref{eq:minMainOP}. $\blacksquare$

The advantage of \eqref{eq:MainOpt}, apart from being convex and enjoying computationally efficient solutions, is that one can incorporate other convex constraints into it for different scenarios. 
In the next sections, we examine this method for fooling neural networks.

\begin{remark}
There are various hypothesis about the nature of adversarial images (see \cite{akhtar_threat_2018}). A popular hypothesis is so-called linearity hypothesis according to which the neural networks are intentionally designed to operate in linear regimes and that makes them susceptible to adversarial examples. The above formulation of the problem basically presupposes that the behavior of \ac{DNN} classifiers around particular image can be approximated by a linear classifier. In this sense, the current formulation is compatible with the linearity hypothesis. 
\end{remark}

Note that the introduced method in \eqref{eq:MainOpt} can also be used for other target functions or learning problems. One can use the cost function used for training as in \cite{goodfellow_explaining_2014} in which case the solution of  \eqref{eq:MainOpt} with $p=\infty$ recovers the adversarial perturbations obtained via the \mbox{\ac{FGSM}}. Again the algorithm can be also used to generate adversarial examples for regression problems. The feasibility problem of \eqref{eq:originalOPII} can be also simplified to 
\begin{equation}\label{eq:MainOptII}
\min_{\bm{\eta}} \| \bm{\eta} \|_{p} \quad \mathrm{s.t.} \quad L(\mathbf{x}) + \bm{\eta}^{\mathrm{T}} \nabla_{\mathbf{x}} L(\mathbf{x})    \leq 0 \, ,
\end{equation}
which recovers the result in \cite{DeepFool} although without the iterative procedure. However the iterative procedure can be easily adapted to the current formulation by repeating the optimization problem until the classifier output changes. In any case, the formulation in  \eqref{eq:originalOPII} provides a general framework for generating adversarial examples using a computationally efficient way.


\section{From Classification to Regression and Other Problems}\label{sec:other}

In this paper we have modeled the generation of adversarial attacks using a convex optimization problem. While we have focused on the task of classification, the convex formulation from \eqref{eq:MainOpt} is not restricted to that specific task. Furthermore, in this section we discuss the applicability of this framework for tasks beside classification. As an example, we apply this framework in the particular context of regression.

In context of regression problems, we assume that the aim of the adversarial perturbation algorithm is to maximize the $\ell_2$-norm of the output perturbation, that is to maximize $L(\mathbf x + \bm \eta) = \| f(\mathbf x) - f(\mathbf x + \bm \eta) \|_2 $ subject to $\| \bm \eta \|_p \leq \epsilon$. In this case finding the adversarial perturbation is indeed solving 
\begin{align}
\argmax_{\bm \eta} \left\lbrace \| \ma{J}_f(\mathbf x)\cdot\bm \eta  \|_{2}^{2} \right\rbrace \quad \mathrm{s.t.} \quad \|\bm \eta \|_p \leq \epsilon \, .
\label{eq:regressionprob}
\end{align}
In this problem, the objective function is quadratic with a positive semi-definite kernel and hence convex. The constraint is also convex. Maximizing convex functions is in general very difficult, however the problem can be solved efficiently in some cases.  For general $p$, the maximum value is related to the operator norm of  $\ma{J}_f(\mathbf x)$. The operator norm of a matrix $\ma A\in\mathbb{C}^{m\times n}$ between $\ell_p$ and $\ell_q$ is defined as \cite{foucart_mathematical_2013}
\[
\|\ma A\|_{p\to q} \triangleq\sup_{\|\ma x\|_p\leq 1}{\|\ma A\ma x \|_q}.
\]
Using this notion, we can see that first $\|\frac{\bm \eta}{\epsilon}  \|_{p}\leq 1$ and therefore
\[
 \| \ma{J}_f(\mathbf x)\cdot\bm \eta  \|_{2}=\epsilon \| \ma{J}_f(\mathbf x)\cdot\frac{\bm \eta}{\epsilon}  \|_{2}\leq \epsilon \|\ma{J}_f(\mathbf x)\|_{p\to 2}.
\]
Therefore the problem of finding a solution to \eqref{eq:regressionprob} amounts to finding the operator norm  $\|\ma{J}_f(\mathbf x)\|_{p\to 2}$. First observe that the maximum value is achieved on the border namely for  $\|\bm \eta \|_p =\epsilon$. In the case of $p=2$, this problem has a closed-form solution.  If $\ma v_{\max}$ is the unit $\ell_2$-norm eigenvector corresponding to the maximum eigenvalue of $\ma{J}_f(\mathbf x)^{\mathrm{T}} \ma{J}_f(\mathbf x)$, then  $\bm \eta=\epsilon\ma v_{\max}$ solves the optimization problem. Note that, the maximum eigenvalue of $\ma{J}_f(\mathbf x)^{\mathrm{T}} \ma{J}_f(\mathbf x)$ corresponds to the square of the  spectral norm $\|\ma{J}_f(\mathbf x)\|_{2\to 2}$.

Another interesting case is when $p=1$, that is when the $\ell_1$-norm of the perturbation is bounded by $\epsilon$. Note that penalizing high $\ell_1$-norm values is a technique used to promote sparsity. When the solution of a problem should satisfy a sparsity constraint, the direct introduction of such constraint into the optimization leads to NP-hardness of the problem. Instead the constraint is relaxed by adding $\ell_1$-norm regularization. The adversarial perturbation designed in this way tends to have only a few non-zero entries. This corresponds to  scenarios like single pixel attacks where only a few pixels are supposed to change. For this choice, we have
\[
 \|\ma A\|_{1\to 2}=\max_{k\in[n]} \|\ma a_k\|_2,
\]
where $\ma a_k$'s are the columns of $\ma A$. Therefore, if the columns of the Jacobian matrix are given by $\ma{J}_f(\mathbf x)=[\ma J_1 \hdots \ma J_{m_0}] $, then
\[
  \| \ma{J}_f(\mathbf x)\cdot\bm \eta  \|_{2}\leq \epsilon \max_{k\in[m_0]}\|\ma J_k\|_2,
\]
and the maximum attained for
\[
\bm\eta^*= \epsilon \ma e_{k^*}\quad\text{ for }\quad k^*=\arg\max_{k\in[m_0]}\|\ma J_k\|_2,
\]
where the vector $\ma e_i$ is the $i$-th canonical vector. This constitutes a single pixel attack.

Finally, the case where the adversarial perturbation is bounded in $\ell_\infty$-norm is of particular interest. 
This bound guarantees that the noise entries have bounded values. 
The problem of finding an adversarial noise corresponds to obtaining the vector for which the operator norm $ \|\ma{J}_f(\mathbf x)\|_{\infty\to 2}$ is attained. Unfortunately this problem turns out to be NP-hard \cite{rohn_computing_2000}. However it is possible to approximately find this norm using semi-definite programming as proposed in \cite{hartman_tight_2015}. The problem is that the semi-definite programming scales badly with input dimension in terms of computational complexity and therefore might not be suitable for fast generation of adversarial examples when the input dimension is very high.

Apart from regression, another example where the above method might be useful to generate adversarial images is the image segmentation problem where there is a class assigned to every pixel of an image. This problem was considered in \cite{segmentation} where the objective of an attacker is to draw certain geometric figures on the output segmentation. In this setup, the noise is designed such that an input is missclassified as certain target class $t$. This constitutes a variation in the type of loss considered in \eqref{eq:L}. Instead of just changing  the output classifier, we aim at changing the output of the classifier into a designated class. In this case, one can instead use $L_t(\mathbf x  + \bm \eta) = f_{k(\mathbf x)} (\mathbf x + \bm \eta) - f_t(\mathbf x + \bm \eta)$, where we have a fixed class $t$ as target. The above analysis applies directly to this problem as well.

Finally, in the context of anomaly detection and monitoring, the goal of an attacker is to maximize the false positives and/or false negatives. This naturally leads to algorithms of the same nature as Algorithm 2 (introduced later on Section \ref{sec:expres}), where a single score function (e.g. the probability of being detected) is the subject of minimization or maximization.

\section{Experiments}\label{sec:expres}
\begin{figure}[bt]
	\centering
	\begin{minipage}[b]{0.85\linewidth}
	\centering
	\begin{tabular}{cc||cc}
	\textbf{Original} & \textbf{Adversarial} & \textbf{Original} & \textbf{Adversarial}  \\
	\hline
	\hline
	\begin{minipage}[b]{0.16\linewidth}
	  \includegraphics[width=0.99\linewidth]{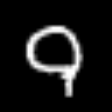}
	\end{minipage}
	&
	\begin{minipage}[b]{0.16\linewidth}
	  \includegraphics[width=0.99\linewidth]{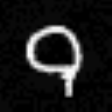}
	\end{minipage}
	&
	\begin{minipage}[b]{0.16\linewidth}
	    \includegraphics[width=0.99\linewidth]{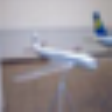}
	\end{minipage}
	&
	\begin{minipage}[b]{0.16\linewidth}
	    \includegraphics[width=0.99\linewidth]{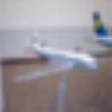}
	\end{minipage}
	\\
	nine &zero & airplane& ship 
	\\
	\hline
	\begin{minipage}[b]{0.16\linewidth}
	  \includegraphics[width=0.99\linewidth]{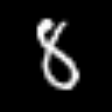}
	\end{minipage}
	&
	\begin{minipage}[b]{0.16\linewidth}
	  \includegraphics[width=0.99\linewidth]{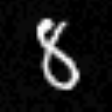}
	\end{minipage}
	&
	\begin{minipage}[b]{0.16\linewidth}
	    \includegraphics[width=0.99\linewidth]{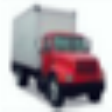}
	\end{minipage}
	&
	\begin{minipage}[b]{0.16\linewidth}
	    \includegraphics[width=0.99\linewidth]{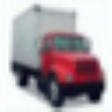}
	\end{minipage}
	\\
	eight &three & truck& car 
	\\
	\hline
	\begin{minipage}[b]{0.16\linewidth}
	  \includegraphics[width=0.99\linewidth]{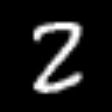}
	\end{minipage}
	&
	\begin{minipage}[b]{0.16\linewidth}
	  \includegraphics[width=0.99\linewidth]{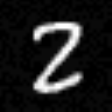}
	\end{minipage}
	&
	\begin{minipage}[b]{0.16\linewidth}
	    \includegraphics[width=0.99\linewidth]{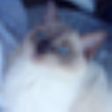}
	\end{minipage}
	&
	\begin{minipage}[b]{0.16\linewidth}
	    \includegraphics[width=0.99\linewidth]{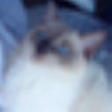}
	\end{minipage}
	\\
	two &three & cat& dog 
	\\
	\hline
	\multicolumn{2}{c}{MNIST dataset} &\multicolumn{2}{c}{CIFAR-10 dataset}
	\end{tabular}
	\end{minipage}
    \caption{Examples of correctly classified images that are missclassfied when adversarial noise is added using Algorithm 1.}
    \label{fig:examples}
\end{figure}
In this section, the Gradient-based Norm-constrained method is used to fool the classifier trained on the task of classification for the MNIST \cite{mnist} and CIFAR-10 \cite{cifar10} datasets. As discussed in Section \ref{sec:avd_and_rub}, for this context of image classification the appropriate loss function $L(\mathbf x)$ to be used in \eqref{eq:MainOpt} is given by \eqref{eq:L}. For this problem, $\| \bm{\eta} \|_{\infty} \leq \epsilon$ is a common constraint that models the undetectability, for sufficiently small $\epsilon$, of adversarial noise by an observer. However solving \eqref{eq:MainOpt} involves finding the function $L(\mathbf x)$ which is defined as the minimum of $m_L-1$ functions with $m_L$ being the number of different classes. In large problems, this may significantly increase the computations required to fool one image. Therefore, we include a simplified version of this algorithm in our simulations. 
The non-iterative methods might not guarantee the fooling of the underlying network but on the other hand, the iterative methods might suffer from convergence problems. 

%

%
\begin{figure}[t]
	\centering
	\begin{minipage}[b]{.48\linewidth}
		\centering
		\centerline{\includegraphics[width=.99\linewidth]{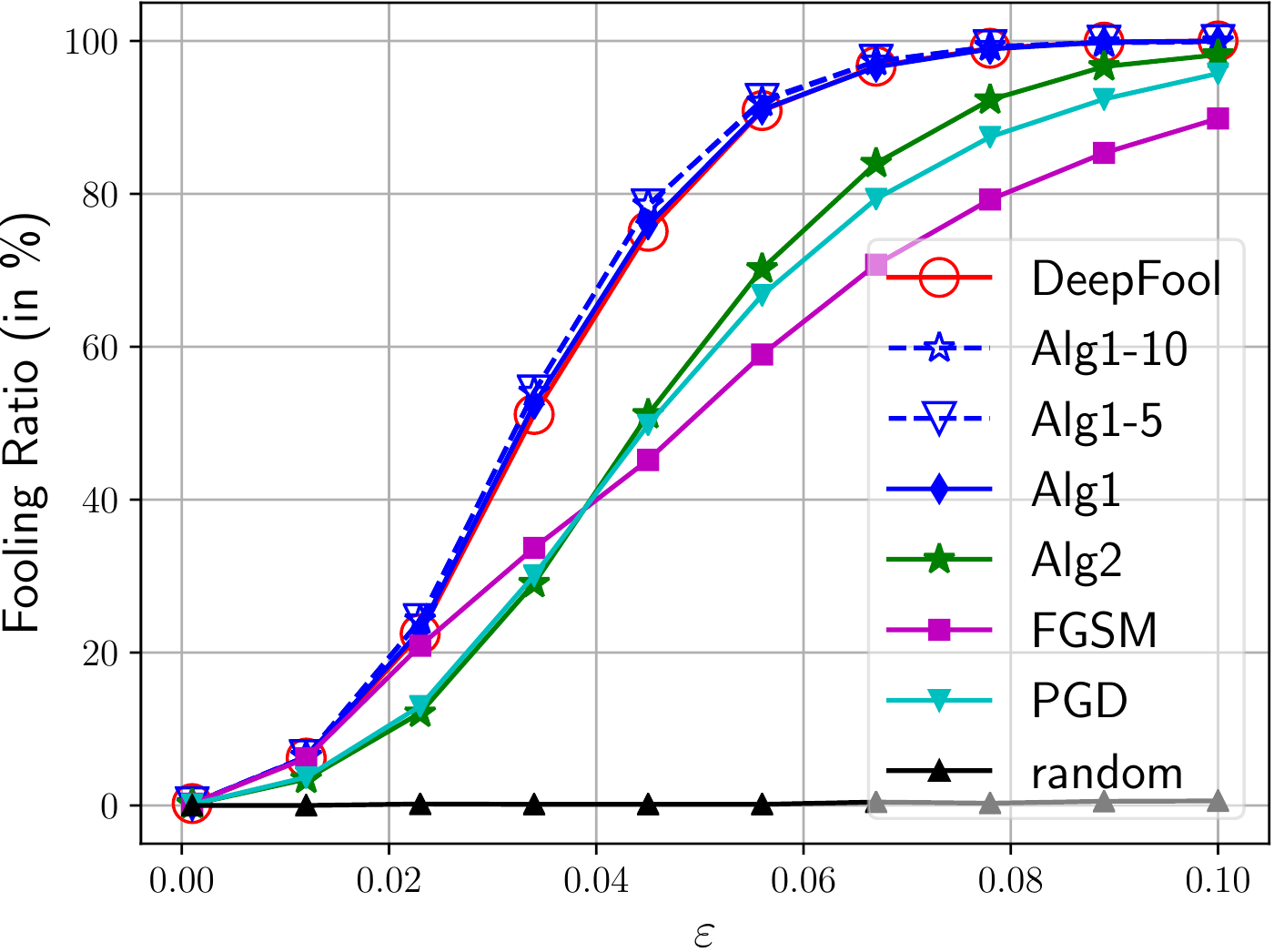}}
		\centerline{(a) FCNN}
	\end{minipage}
	~
	\begin{minipage}[b]{.48\linewidth}
		\centering
		\centerline{\includegraphics[width=.99\linewidth]{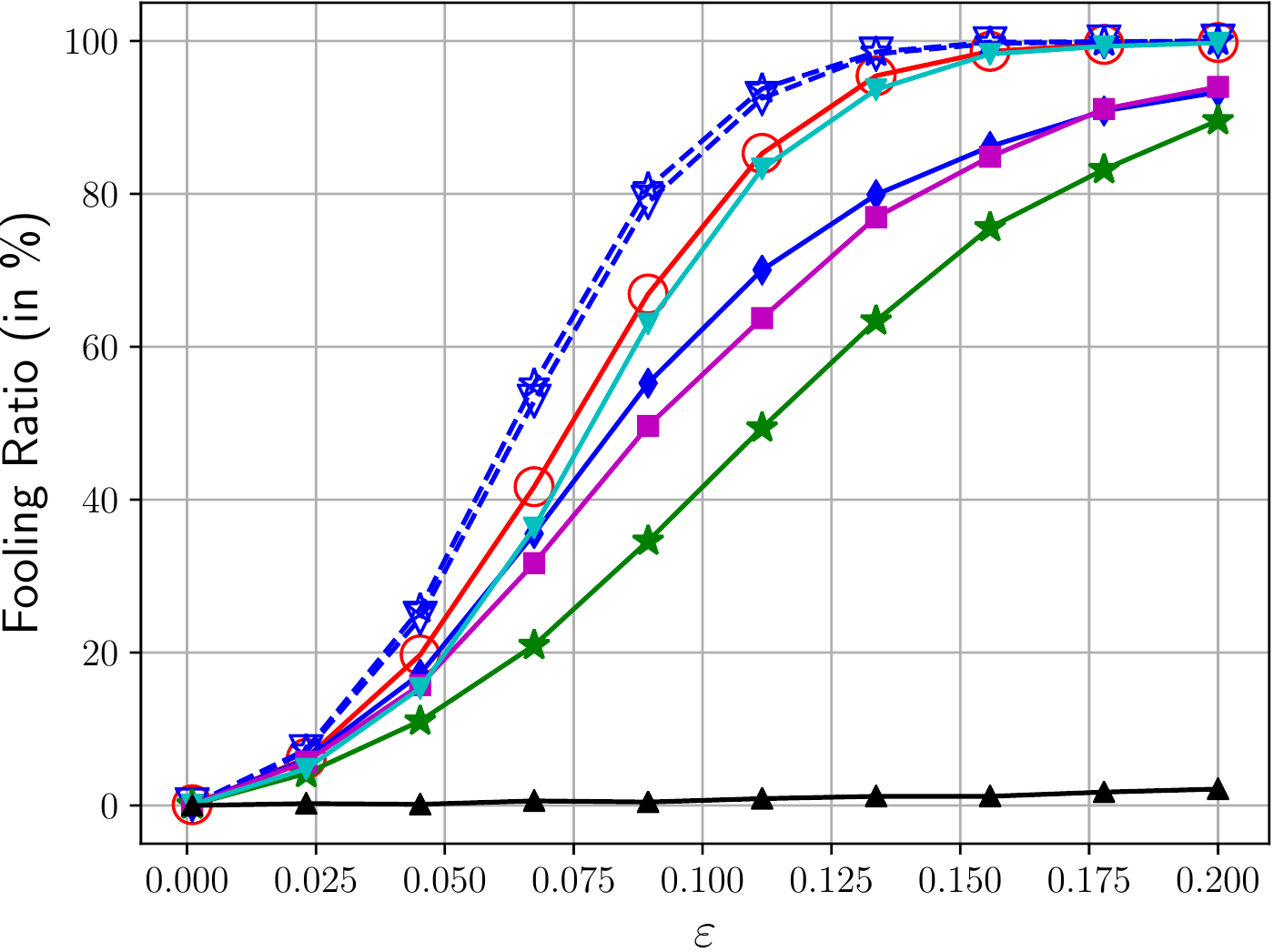}}
		\centerline{(b) LeNet-5}
	\end{minipage}

	\begin{minipage}[b]{.99\linewidth}
		\vertfill{$\,$}
	\end{minipage}

	\begin{minipage}[b]{.48\linewidth}
		\centering
		\centerline{\includegraphics[width=.99\linewidth]{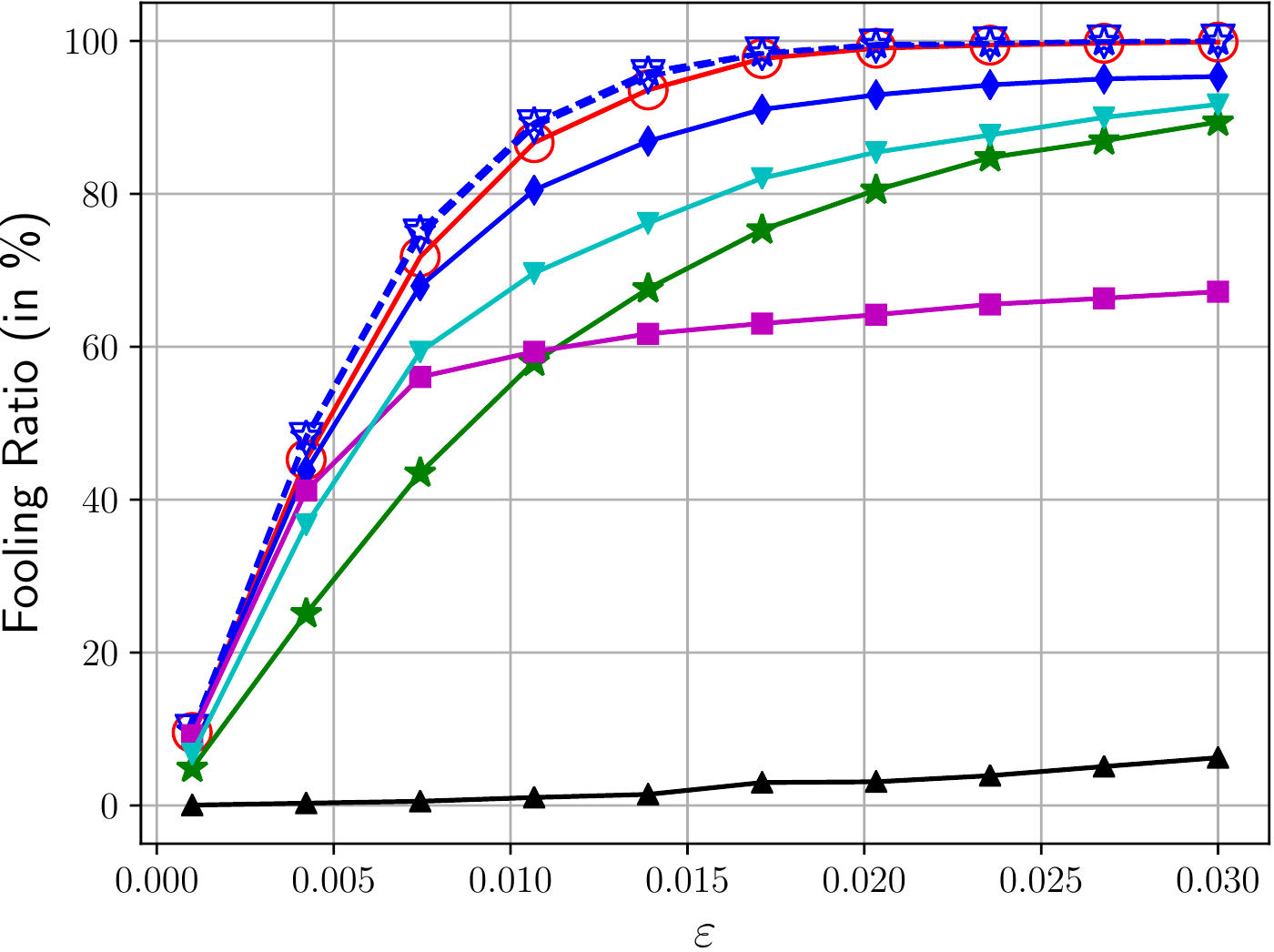}}
		\centerline{(c) NIN}
	\end{minipage}
	~
	\begin{minipage}[b]{.48\linewidth}
		\centering
		\centerline{\includegraphics[width=.99\linewidth]{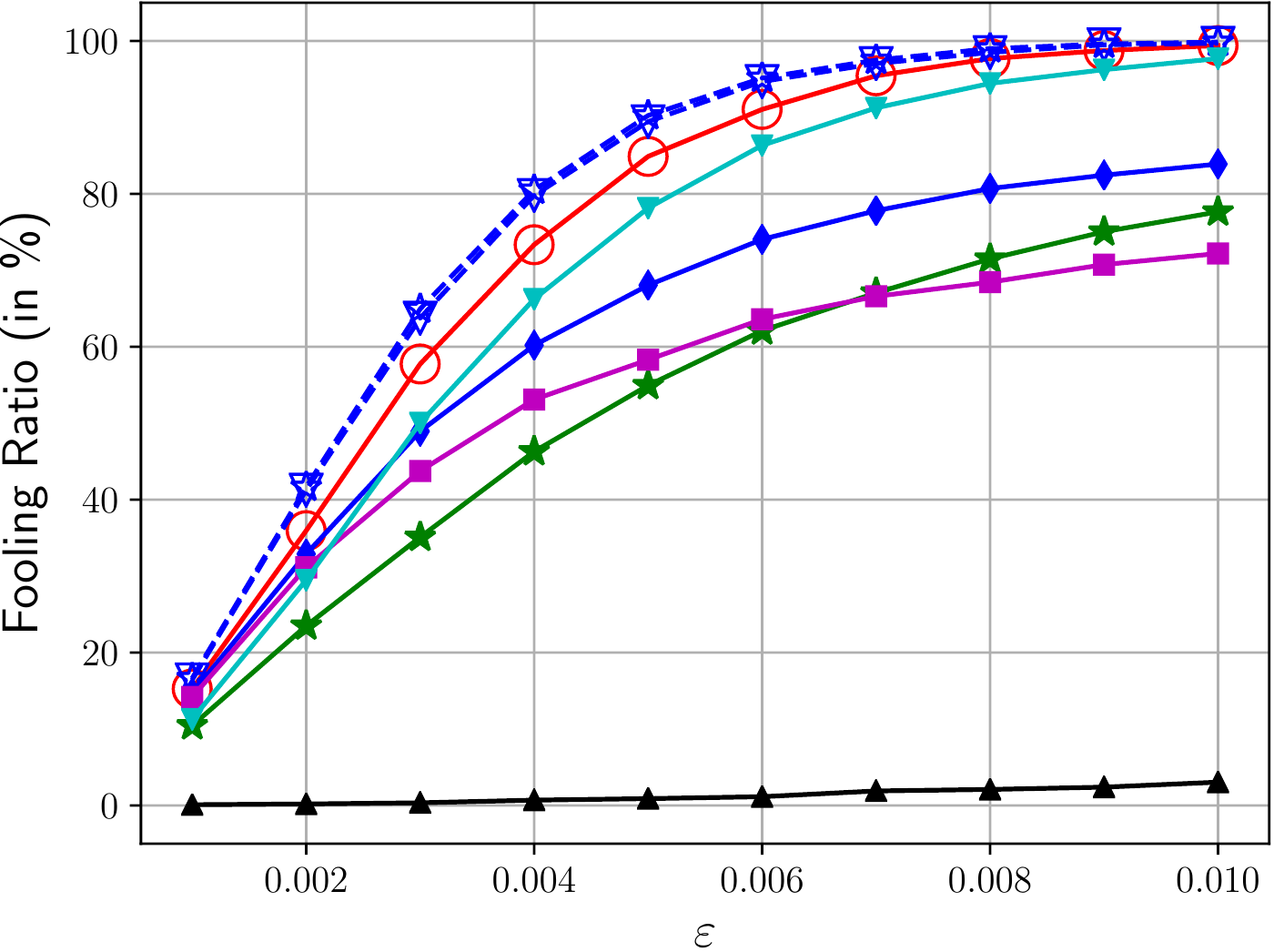}}
		\centerline{(d) DenseNet}
	\end{minipage}
	\caption{(a) and (b): Fooling ratio of the adversarial samples for different values of $\epsilon$ on the MNIST test dataset. (c) and (d): Fooling ratio of the adversarial samples for different values of $\epsilon$ on the CIFAR-10 test datasets.}
	\label{fig:res}
\end{figure}

To benchmark the proposed adversarial algorithms, we consider following methods tested on the aforementioned datasets:
\begin{itemize}
\item \textbf{Algorithm 1}: This algorithm solves \eqref{eq:MainOpt} with $L(\mathbf x)$ given by \eqref{eq:L}. Note that, for evaluating $L$ at a given $\mathbf x$ one must search over all $l  \neq k(\mathbf x)$. This can be computationally expensive when the number of possible classes (i.e., the number of possible values for $l$) is large. The $\ell_\infty$-norm is chosen for the constraint. Moreover, an example of adversarial images obtained using this algorithm is shown in Figure \ref{fig:examples}.
\item \textbf{Algorithm 1}-$n$: Iterative version of Algorithm 1 with $n$ iterations. The adversarial noise is the sum of $n$ noise vectors with $\ell_\infty$-norm of $\epsilon / n$, computed through $n$ successive approximations.
\item \textbf{Algorithm 2}: This algorithm approximates \eqref{eq:L} with $L(\mathbf x) \approx f_{k(\mathbf{x})}(\mathbf{x})$, thus reducing the computation of $L(\mathbf x)$ when the number of classes is large. Note that we cannot use $L(\mathbf x + \bm{\eta})<0$ to guarantee that we have fooled the network. Nevertheless, the lower the value of $L(\mathbf x + \bm{\eta})$ the most likely it is that the network has been fooled. The same reasoning is valid for the \ac{FGSM}.
\item \textbf{\ac{FGSM}}: This well-known method was proposed by \cite{goodfellow_explaining_2014} where $L$ is replaced by the negative training loss. Usually the cross-entropy loss is used for this purpose. With the newly replaced function,  \eqref{eq:MainOpt} is solved for $p=\infty$.
\item \textbf{DeepFool}: This method was designed by \cite{DeepFool} and makes use of iterative approximations. Every iteration of DeepFool can be written within our framework by replacing $L$ by 
\begin{align*}
L(\mathbf x + \bm{\eta} ) &= f_{k(\mathbf{x})}(\mathbf{x} + \bm{\eta}) - f_{\hat l}(\mathbf{x} + \bm{\eta})
\, , \\ \text{where} \qquad
\hat l &= \argmin_{l \neq k(\mathbf x)} \left\{ \frac{|f_{k(\mathbf{x})}(\mathbf{x}) - f_l(\mathbf{x})|}{ \|\nabla f_{k(\mathbf{x})}(\mathbf{x}) - \nabla f_l(\mathbf{x})\|_q } \right\} \, .
\end{align*}
The adversarial perturbations are computed using $p = \infty$, thus $q=1$, with a maximum of $50$ iterations. These parameters were taken from \cite{DeepFool}.
\item \textbf{PGD}: This method is an iterative version of the \ac{FGSM} where the initial point is randomly chosen from an $\epsilon$ vicinity of $\mbf x$ \cite{pgd_attack}.
\item \textbf{Random}: For benchmarking purpose, we also consider random noise with independent Bernoulli distributed entries with $\mathbb P (X = \epsilon) = \mathbb P (X =-\epsilon) = \frac{1}{2} $.
\end{itemize}

The above methods are tested  on the following deep neural network architectures:
\begin{itemize}
  \item \textbf{MNIST} : A fully connected network with two hidden layers of size $150$ and $100$ respectively, as well as the LeNet-$5$ architecture \cite{lenet}. 
  \item \textbf{CIFAR-10} : The Network In Network (NIN) architecture \cite{nin}, and a $40$ layer DenseNet \cite{densenet}.
\end{itemize}

As a performance measure, we use the \emph{fooling ratio} defined in \cite{DeepFool} as the percentage of correctly classified images that are missclassified when adversarial perturbations are applied. Of course, the fooling ratio depends on the constraint on the norm of adversarial examples. 
Therefore, in Figure \ref{fig:res} we observe the fooling ratio for different values of $\epsilon$ on the aforementioned neural networks. 
As expected, the increased computational complexity of iterative methods such as DeepFool and Algorithm 1-$n$ translates into increased performance with respect to non-iterative methods. 
Nevertheless, as shown in Figures \ref{fig:res}(a) and (c), the performance gap between iterative and non-iterative algorithms is not always significant. 
For the case of iterative algorithms, the proposed Algorithm 1-$n$ outperforms DeepFool. The same holds true for Algorithm 1 with respect to other non-iterative methods such as the \ac{FGSM}, while Algorithm 2 obtains competitive performance with respect to the \ac{FGSM}. 

Finally, we measure the robustness of different networks using $\hat{\rho}_1(f)$ and $\hat{\rho}_2(f)$, with $p=\infty$. We also include the minimum $\epsilon$, such that DeepFool obtains a fooling ratio greater than 99\%, as a performance measure as well. These results are summarized in Table \ref{tab:robust}, where we obtain coherent results between the $3$ measures.

\begin{table}[htb]
    \centering
    \begin{tabular}{l|c|c|c|c}
    & \small{Test} & \small{$\hat{\rho}_1(f)$} & \small{$\hat{\rho}_2(f)$}  & \small{fooled} \\
    & \small{error} &\small{\cite{DeepFool}} & \small{(ours)}  & \small{$>$99\%} \\
    \hline
    \small{FCNN (MNIST)}& \small{1.7\%} & 0.036 & 0.034  & $\epsilon =$0.076 \\
    \small{LeNet-5 (MNIST)}& \small{0.9\%} & \textbf{0.077} & \textbf{0.061}  & $\epsilon =$\textbf{0.164} \\
    \hline
    \small{NIN (CIFAR-10)}& \small{13.8\%} & \textbf{0.012} & \textbf{0.004}  & $\epsilon =$\textbf{0.018} \\
    \small{DenseNet (CIFAR-10)}& \small{5.2\%} & 0.006 & 0.002  & $\epsilon =$0.010
    \end{tabular}
    \caption{Robustness measures for different classifiers.}
    \label{tab:robust}
\end{table}
 
\section{Conclusion}
In this paper, we have shown that the perturbation analysis of different models leads to methods for generating adversarial examples via convex programming. For classification we have formulated already existing methods as special cases of the proposed framework. Moreover, novel methods for designing adversarial noise under various desirable constraints have been derived. Finally the applicability of this framework has been tested for classification through empirical simulations of the fooling ratio, benchmarked against the well-known \ac{FGSM}, \ac{PGD} and DeepFool methods. We have also discussed how the current framework can be extended to variety of different problems. As future works, it is still worth exploring the reason behind the existence of adversarial examples, and the design of effective defenses.

\clearpage
\bibliographystyle{IEEEbib}
\bibliography{main}

\clearpage

\end{document}